\documentclass[onecolumn,authoryear]{els-mrw} 

\usepackage{amsmath,amssymb,amsfonts,amsthm,makeidx,graphicx}
\usepackage{txfonts}
\usepackage{helvet}
\usepackage{multirow}
\usepackage{amsthm}
\usepackage{algorithm}
\usepackage{algorithmic}
\usepackage{subfigure}
\usepackage{makecell}
\graphicspath{ {figure/} }

\begin{document}

\chapter{Privacy-Preserving Distributed Optimization and Learning}\label{chap1}

\author[1]{Ziqin Chen}%
\author[1,2]{Yongqiang Wang}%

\address[1]{\orgname{Clemson University}, \orgdiv{Department of Electrical and Computer Engineering}, \orgaddress{Clemson, SC 29634 USA (e-mail: ziqinc@clemson.edu; yongqiw@clemson.edu)}}
\address[2]{\orgname{Corresponding Author}}

\articletag{Optimization and Multi-Agent Systems}

\maketitle

\begin{abstract}[Abstract]
Distributed optimization and learning has recently garnered great attention due to its wide applications in sensor networks, smart grids, machine learning, and so forth. Despite rapid development, existing distributed optimization and learning algorithms require each agent to exchange messages with its neighbors, which may expose sensitive information and raise significant privacy concerns. In this survey paper, we overview privacy-preserving distributed optimization and learning methods. We first discuss cryptography, differential privacy, and other techniques that can be used for privacy preservation and indicate their pros and cons for privacy protection in distributed optimization and learning. We believe that among these approaches, differential privacy is most promising due to its low computational and communication complexities, which are extremely appealing for modern learning based applications with high dimensions of optimization variables. We then introduce several differential-privacy algorithms that can simultaneously ensure privacy and optimization accuracy. Moreover, we provide example applications in several machine learning problems to confirm the real-world effectiveness of these algorithms. Finally, we highlight some challenges in this research domain and discuss future directions. 
\end{abstract}
\begin{keywords}
Distributed optimization and learning, differential privacy, homomorphic cryptography, privacy preservation, secure multi-party computation.
\end{keywords}

\section{Introduction}\label{section1}
In recent years, the rapid development of large-scale networks and big data has led to the widespread applications of distributed optimization and learning. In this paradigm, each agent has a private objective function and engages in communicating with neighboring agents to cooperatively learn an optimal solution to a global objective. Due to its inherent advantages in scalability and privacy, distributed optimization/learning methods have found extensive applications in various fields, including sensor networks, smart grids, formation control, machine learning and so on~\citep{yang2019survey,machinelearning}. Traditional distributed optimization/learning methods are centered around batch or offline learning, that is, the algorithm is trained by using a dataset acquired before implementing the algorithm, which limits their applicability in numerous practical scenarios where data are acquired in a serial manner. Recognizing this limitation, online optimization and learning has emerged as an active research field in the past two decades. Online optimization/learning allows for the sequential access and processing of data, making them particularly appealing for large-scale datasets and dynamic scenarios where data is continually generated, such as social media streams and real-time sensor interpretation~\citep{li2023}.

Although significant progress has been made in both distributed offline and online optimization/learning, all of existing results require agents to share messages (learned parameters or gradients) in each iteration, which will pose privacy concerns, especially when the training dataset is proprietary to each agent and contains sensitive information, such as medical or financial records, web search history, and more~\citep{financial2,legal1,legal2}.
In fact, recent works~\cite{huang2015},~\cite{zhang2018}, and~\cite{infer} have shown that without a strong privacy mechanism in place, external adversaries can easily reconstruct individuals’ raw data from shared messages. Therefore, developing privacy-preserving algorithms for distributed optimization and learning is crucial. Along this line, plenty of privacy-preserving approaches have been reported to address potential privacy breaches in distributed optimization/learning. One approach involves secure multi-party computation, like secret sharing and homomorphic encryption~\citep{MPCzongshu,zhang2018,zhang2019}. However, these approaches often come with significant communication and computational overheads. Moreover, except our prior works~\cite{zhang2018} and~\cite{zhang2019}, most existing secure multi-party computation results rely on a ``centralized" data aggregator, which does not exist in the fully distributed setting. Another approach capitalizes on
the ``structure" properties of distribution optimization to inject temporally or spatially correlated uncertainties for privacy, as in~\cite{yan2013},~\cite{lou2017}, and our prior works~\citep{zhang2018privacy,Wang2022quantization,gaohuan2023,uncoordinated2023}. However, the injection of correlated uncertainties results in the privacy strength of these approaches being inherently limited by the optimization problems' intrinsic properties. Differential privacy (DP) has achieved remarkable success and has become a de facto standard for privacy protection in recent years. Nevertheless, most DP results in distributed optimization/learning face a dilemma of trading optimization accuracy for privacy, which significantly impedes its further development, especially in accuracy-sensitive applications. Our recent results~\citep{Wangtailoring2023,NEseeking2022,Aggregative2024,Chengradient2023,Chengradienttracking2023} have successfully circumvented this dilemma, ensuring rigorous DP and optimization accuracy simultaneously.

This paper aims to provide a survey of privacy-preserving methods for distributed optimization and learning. It is structured around four perspectives: literature review, backgrounds, algorithms, and example applications. Although the survey papers by~\cite{survey1} and \cite{antwi2021} have explored the intersection of privacy and collaborative deep learning, our investigation provides a more comprehensive review and new perspectives. More specifically,~\cite{survey1} provided a generic review but lacks an in-depth focus on privacy preservation in distributed optimization and learning, including fields like noncooperative games and distributed online learning.~\cite{antwi2021} was primarily concentrated on the homomorphic encryption method, a focus markedly distinct from our objectives. Our contribution is a detailed overview of existing privacy-preserving methods, with a special emphasis on differential-privacy algorithms that are capable of  ensuring both privacy and optimization accuracy. By providing this review, we aim not only to fill the gap identified in previous surveys but also to inspire further research in this field. 

\section{Literature Review}\label{section2}
In this section, we provide a review of the commonly used privacy-preserving approaches in distributed optimization and learning, including homomorphic encryption, secure multi-party computation, differential privacy, and various other methods aimed at ensuring data confidentiality. Relevant literature on these approaches is briefly summarized in Table~\ref{table1}.
%\footnotemark{a}
\begin{table}[H]
\small
\TBL{\caption{Commonly used privacy-preserving approaches in distributed optimization and learning.}\label{table1}}
{\begin{tabular*}{\textwidth}{@{\extracolsep{\fill}}@{}lp{7cm}p{4cm}@{}}
\toprule
\multicolumn{1}{@{}l}{\TCH{Privacy-preserving methods}} &
\multicolumn{1}{c}{\TCH{Relevant literature}}
& \multicolumn{1}{c}{\TCH{Drawbacks}}  \\
\colrule
\multirow{1}*{Homomorphic cryptography}  &
Distributed offline optimization:~\citep{shoukry2016,lu2018,Tang2019,alexandru2020,cheng2021,zhang2021privacy,yan2021,wu2021privacy,chen2022,Huo2022,zhang2018,zhang2019}.

Noncooperation game:~\citep{lu2015b}.

Distributed online learning:~\citep{wang2019privacy}.
& Heavy computational and communicational overheads;

Specific computation-types limitations in secret sharing.\\
\colrule
\multirow{1}*{Secure multi-party computation}  & 

Distributed offline optimization:~\citep{wagh2020,huo2022distributed,xie2022efficient,tian2023fully}.

Noncooperation game:~\citep{abraham2006distributed,zhang2013rational}.

Distributed online learning:~\citep{dong2020privacy}.
 & Heavy computational and communicational overheads\\
\colrule
\multirow{1}*{Differential privacy}  & Distributed offline optimization:~\citep{huang2015,nozari2016,han2016,zhang2016,cloud2016,hale2017,zhang2018improving,zhang2019,huang2019,ding2021,chen2023,xuan2023,Wangtailoring2023,nonconvex2023,Shi2024}.

Noncooperation game:~\citep{gade2020aggregate,ye2021aggregative,NEseeking2022,WangJimin2022,Aggregative2024}. 

Distributed online learning:~\citep{zhu2018,li2018,hou2019online,xiong2020,hu2021online,Lu2021,han2022online,liu2022online,chen2023online,Lu2023online,Yuan2023,cheng2023online,Chengradient2023,Chengradienttracking2023,zhao2024online}.
& The tradeoff between privacy and optimization accuracy\\
\botrule
\end{tabular*}}{}
\end{table}
\subsection{Homomorphic encryption}
Homomorphic encryption was first proposed by~\cite{Rivest1978} and continuously developed over the past three decades~\citep{marcolla2022survey,doan2023survey}. This method enables certain algebraic operations on ciphertexts to produce an encrypted result, which, after decryption, matches the results of operations performed on plaintexts. According to the types of computations supported by homomorphic encryption, it can be classified into partially and fully homomorphic encryption. Partially homomorphic encryption allows the specific computation (e.g., addition or multiplication) on encrypted data, whereas fully homomorphic encryption supports arbitrary computations (e.g., both addition and multiplication operations). Recently, partially homomorphic encryption have been applied in distributed optimization/learning (see Table~\ref{table1}). For example,~\cite{lu2015b} developed a distributed Nash equilibrium seeking algorithm using reinforcement learning and homomorphic encryption, achieving convergence to a Nash equilibrium for discrete constrained potential games.~\cite{shoukry2016} and~\cite{alexandru2020} introduced privacy-preserving protocols relying on partially homomorphic encryption for quadratic program problems. However, all these results require a trusted cloud for computation, making them inapplicable to the completely distributed setting. Similar limitations are observed in~\cite{lu2018},~\cite{Tang2019}, and~\cite{cheng2021}. Only
our prior homomorphic-encryption-based results~\citep{zhang2018,zhang2019} can achieve both privacy and optimal accuracy without relying on any aggregator or third party. In addition, since distributed homomorphic encryption requires agent interaction and local computation performed on encrypted data, as the number of participating agents grows, both communication and computational complexities will significantly increase. In fact, homomorphic encryption often results in an exponential growth in ciphertext sizes,which is often far exceeding the size of the original plaintext. Hence, distributed homomorphic encryption methods demand a large amount of computational and communication resources, presenting significant challenges for development in large-scale machine learning applications.

\subsection{Secure multi-party computation}
Secure multi-party computation (MPC) was first introduced in~\cite{yao1982}. It aims to design a secure protocol that enables multiple participants $P_{i},~i=1,\cdots,m$ to collaboratively compute an objective function $f(x_{1},\cdots,x_{m})=(y_{1},\cdots,y_{m})$ using their private inputs $x_{i}$, while ensuring each participant $P_{i}$ receives only its own corresponding output $y_{i}$ with no additional information, thereby preserving privacy. This concept has evolved to include various protocols, such as garbled circuit, secret sharing, oblivious transfer and so on~\citep{MPCzongshu}. Traditional MPC protocols are often designed for the two-party scenario~\citep{2MPC1,2MPC3}. For scenarios involving more than two parties, algorithms based on three-party and multi-party communication have also been developed~\citep{3MPC2,3MPC3}. Recently, secret sharing, noted for its simplicity and interactivity, has been applied in distributed optimization and learning. For example,~\cite{wagh2020} utilized secret sharing to protect customer privacy in distributed smart grids.~\cite{huo2022distributed} introduced a privacy-preserving electric vehicle charging algorithm by using Shamir's secret sharing to ensure user privacy.~\cite{tian2023fully} implemented secret sharing in fully distributed privacy-preserving optimization, showing its efficacy in protecting sensitive information. Although enabling participant agents to collaborate without requiring a trusted party, distributed MPC still requires a certain level of trust among participants (for example, secret sharing needs collaboration from a threshold number of participants to reconstruct the secret). In addition, the reliance on computing the objective function $f(x_{1},\cdots,x_{m})$ in MPC indicates that increasing participant-agent numbers will also increase both computational and communication complexities, leading to a challenge in scalability. While secret sharing can reduce the privacy-preserving-computation cost, its suitability is limited to specific types of computations, potentially restricting its applications in the diverse data processing requirements in distributed optimization and learning.

\subsection{Differential privacy}
Differential Privacy (DP) was first proposed by~\cite{dwork2006}. It is realized by introducing independent noises to perturb the algorithm such that the probability distribution of its output remains relatively insensitive to modifications in any single record of the input~\citep{dwork2014}. DP distinguishes itself from homomorphic encryption and MPC approaches by its low computational and communication demands and its robustness against arbitrary side information. This robustness ensures that DP's efficacy is not significantly compromised by additional information that an adversary may acquire from other sources, a fact supported by~\cite{kasiviswanathan2008}. 

Nowadays, numerous efforts have been made to apply the DP framework into distributed optimization and learning, as elaborated in Table~\ref{table1}. In these works, DP's implementation typically employs two approaches: output perturbation and objective perturbation. Output perturbation requires
solving the optimization problem first and then adding Laplace or Gaussian noise to the output variables. This approach preserves the original objective functions, making the algorithms 
effectively approximate the optimal solution to the original problem.  Objective perturbation, entails adding a noisy term to the objective functions first and then solving the perturbed optimization problem. This approach, unfortunately, is only applicable when the objective function is precisely known to individual agents, which is not the case in most learning applications. A comparison of existing DP approaches in distributed optimization and learning is summarized in Tables~\ref{table2}-\ref{table4}.

Although DP provides a promising paradigm for privacy protection in distributed optimization/learning, directly incorporating persistent DP-noise into existing distributed optimization/learning algorithms will compromise optimization accuracy, leading to a fundamental tradeoff between privacy and accuracy. To the best of our knowledge, most existing DP results for distributed optimization and learning have to face this tradeoff. Typically, most current DP results terminate the algorithm after a pre-determined number of iterations, with this number calculated offline according to the desired privacy budget (privacy level). This approach invariably leads to an optimization error, whose magnitude is inversely proportional to the privacy budget. On another front, some DP results only bound the privacy budget for a single agent in a single iteration~\citep{zhang2016,hale2017,huang2019}. However, given that an adversary could leverage all intermediate outputs for inference, the privacy budget accumulates throughout the iterative process, 
thereby leading to a decaying privacy protection over time. It is worth noting that our recent works~\citep{Wangtailoring2023,NEseeking2022,Aggregative2024,Chengradient2023,Chengradienttracking2023} have successfully circumvented the tradeoff between optimization accuracy and privacy. In these works, we ensure both convergence and rigorous DP with a finite privacy budget, even when the number of iterations tends to infinity.

In addition, some DP results in distributed optimization and learning require a trusted curator for data aggregation and distribution. For example,~\cite{cloud2016} and~\cite{hale2017} rely on a trusted cloud that collects raw data, subsequently adds noise, and then distributes the noised-data to each participant agent. Similarly,~\cite{huang2019} introduced a DP distributed optimization algorithm using the augmented direction method of multipliers, which requires a trusted ``centralized" server to average updated primal variables of all agents in each iteration. Besides these approaches that explicitly require a trusted third party, most of existing DP results in distributed optimization/learning still use the conventional ``centralized" DP framework, which, in the absence of a data aggregator/curator, requires participating agents to trust each other and cooperatively determine the amount of noise needed to achieve a certain level of privacy protection (detailed explanation is given in Subsubsection~\ref{section323}). To implement DP in the fully distributed setting, where an agent does not trust anyone else (including other participating agents) and aims to protect against an adversary that can observe every message shared in the network, the approach of local differential privacy (LDP) has to be introduced~\citep{Chengradient2023,hou2019online,Chengradient2023,Chengradienttracking2023}. In fact, LDP is widely regarded as the strongest framework of differential privacy~\citep{LDPzongshu}. 
\begin{table}
\small
\TBL{\caption{Comparison of differential-privacy approaches in distributed offline optimization}\label{table2}}
{\begin{tabular*}{\textwidth}{@{\extracolsep{\fill}}@{}lllllllll@{}}
\toprule
\multirow{2}*{\TCH{Literature}} 
&\multirow{2}*{\TCH{Privacy}}
& \multirow{1}*{\TCH{Perturbed}}
& \multirow{1}*{\TCH{Privacy budget}}
&\multicolumn{3}{l}{\TCH{Accuracy upper bound}} 
& \multirow{2}*{\TCH{Tradeoff?}} \\
\cline{5-7}
& & \multirow{1}*{\TCH{term}}  & \multirow{1}*{\TCH{characterizion}\footnotemark{a}} & \multicolumn{1}{l}{\TCH{Nonconvex}} & \multicolumn{1}{l}{\TCH{Convex}} & \multicolumn{1}{c}{\TCH{Strongly convex}} &\\
\colrule
\cite{huang2015} & $\epsilon$-DP & Output & $\infty$ & -- & -- & $\mathcal{O}\left(\frac{1}{\epsilon^2}\right)$  & Yes\\
\cite{nozari2016} & $\epsilon$-DP & Objective & $\infty$ & -- &  $\mathcal{O}\left(\frac{1}{\epsilon}\right)$ & -- &  Yes\\
\cite{han2016} & $\epsilon$-DP & Output & $T$ & -- &  $\mathcal{O}\left(\frac{1}{\epsilon^{\frac{1}{4}}}\right)$ & -- &  Yes\\
\cite{zhang2016} & $\epsilon$-DP & Output & $t$ & -- & -- & $\mathcal{O}\left(\frac{1}{\epsilon^2}\right)$ & Yes\\
\cite{cloud2016} & $\epsilon$-DP & Objective & $\infty$ & -- &  $\mathcal{O}\left(\frac{1}{\epsilon^2}\right)$ & -- &  Yes\\
\cite{hale2017} & $\epsilon$-DP & Output & $t$ &-- &  $\mathcal{O}\left(\frac{1}{\epsilon^2}\right)$  & --  & Yes\\
\cite{zhang2018improving} & $\epsilon$-DP & Output & $T$ & -- & -- & $\mathcal{O}\left(\frac{1}{\epsilon^2}\right)$ & Yes\\
\cite{zhang2019} & $\epsilon$-DP & Objective & $T$ & -- &  $\mathcal{O}\left(\frac{1}{\epsilon^2}\right)$ & -- &  Yes\\
\cite{huang2019} & $(\epsilon,\delta)$-DP & Output & $t$ & -- &  $\mathcal{O}\left(\frac{\sqrt{\delta}}{\epsilon^2}\right)$  & --  & Yes\\
\cite{ding2021} & $\epsilon$-DP & Output & $\infty$ & -- & -- & $\mathcal{O}\left(\frac{1}{\epsilon^2}\right)$  & Yes\\
\cite{chen2023} & $\epsilon$-DP & Output & $T$ & -- & -- & $\mathcal{O}\left(\frac{1}{\epsilon}\right)$ & Yes\\
\cite{xuan2023} & $\epsilon$-DP & Output & $T$ & -- & -- & $\mathcal{O}\left(1\right)$ & Yes\\
\cite{Wangtailoring2023} & $\epsilon$-DP & Output & $\infty$ & -- &  0  & -- &  No\\
\cite{nonconvex2023} & $(\epsilon,\delta)$-DP & Objective & $t$ & 0 &  --  & -- &  No\\
\cite{liu2024} & $(\epsilon,\delta)$-DP & Objective & $T$ & -- &  $\mathcal{O}\left(\frac{1}{\epsilon}\right)$   & $\mathcal{O}\left(\frac{1}{\epsilon^2}\right)$  &  Yes\\
\cite{Shi2024} & $\epsilon$-DP & Output & $\infty$ & -- & -- & $\mathcal{O}\left(1\right)$  & Yes\\
\botrule
\end{tabular*}}{\begin{tablenotes}
\footnotetext[a]{We use ``Privacy budget characterization" to represent how the work characterizes a finite privacy budget. Specifically, ``$t$" represents that the work only analyzed the privacy budget in a single iteration (This category includes works that demonstrate a finite privacy budget ``$T\epsilon$" across a finite number of iterations ``$T$", using the composition theorem). ``$T$" implies that the work proved a finite cumulative privacy budget in a finite number of iterations. ``$\infty$" represents that the work can achieve rigorous DP with a finite cumulative privacy budget, even when the number of iterations tends to infinity.}
\end{tablenotes}}
\end{table}
\begin{table}
\small
\TBL{\caption{Comparison of differential-privacy approaches in noncooperative games.}\label{table3}}
{\begin{tabular*}{\textwidth}{@{\extracolsep{\fill}}@{}llllllll@{}}
	\toprule
	\multirow{2}*{\TCH{Literature}} 
	&\multirow{2}*{\TCH{Privacy}}
	& \multirow{1}*{\TCH{Perturbed}}
	& \multirow{1}*{\TCH{Privacy budget}}
	&  \multirow{2}*{\TCH{Game}}
	& \multirow{1}*{\TCH{Accuracy}}
	& \multirow{2}*{\TCH{Tradeoff?}} \\
	& & \multirow{1}*{\TCH{term}}  & \multirow{1}*{\TCH{characterizion}\footnotemark{a}} &  &\multirow{1}*{\TCH{upper bound}}  &\\
\colrule
\cite{ye2021aggregative} & $\epsilon$-DP & Output & $\infty$ & Aggregative game &  $\mathcal{O}(\frac{1}{\epsilon})$ & Yes\\
\cite{WangJimin2022} & $\epsilon$-DP & Output & $t$ & Aggregative game &  $\mathcal{O}(\frac{1}{\epsilon^2})$ & Yes\\
\cite{NEseeking2022} & $\epsilon$-DP & Output & $\infty$ & Normal-form game &  0 & No\\
\cite{Aggregative2024} & $\epsilon$-DP & Output & $\infty$ & Aggregative game &  0 & No\\
\botrule
\end{tabular*}}{}
\end{table}
\begin{table}
\small
\TBL{\caption{Comparison of differential-privacy approaches in distributed online optimization and learning.}\label{table4}}
{\begin{tabular*}{\textwidth}{@{\extracolsep{\fill}}@{}p{5cm}lllllll@{}}
\toprule
\multirow{2}*{\TCH{Literature}} 
&\multirow{2}*{\TCH{Privacy}}
& \multirow{1}*{\TCH{Perturbed}}
& \multirow{1}*{\TCH{Privacy budget}}
&\multicolumn{2}{l}{\TCH{Accuracy upper bound}} 
& \multirow{2}*{\TCH{Tradeoff?}} \\
\cline{5-6}
& & \multirow{1}*{\TCH{term}}  & \multirow{1}*{\TCH{characterizion}\footnotemark{a}}  & \multicolumn{1}{l}{\TCH{Convex}} & \multicolumn{1}{c}{\TCH{Strongly convex}} &\\
\colrule
\cite{zhu2018,xiong2020,Lu2021,chen2023online} & \multirow{2}*{$\epsilon$-DP} & \multirow{2}*{Output} & \multirow{2}*{$t$} & \multirow{2}*{$\mathcal{O}(\frac{1}{\epsilon^2})$}  & \multirow{2}*{$\mathcal{O}(\frac{1}{\epsilon^2})$} & \multirow{2}*{Yes} \\
\colrule
\cite{li2018,hou2019online} & $(\epsilon,\delta)$-DP & Output & $T$ & $\mathcal{O}(\frac{1}{\epsilon^2})$  & $\mathcal{O}(\frac{1}{\epsilon})$ & Yes\\
\colrule
\cite{hu2021online,han2022online,Lu2023online,zhao2024online} & \multirow{2}*{$\epsilon$-DP} & \multirow{2}*{Output} & \multirow{2}*{$t$} & \multirow{2}*{$\mathcal{O}(\frac{1}{\epsilon^2})$}  & \multirow{2}*{--} & \multirow{2}*{Yes}\\
\colrule
\cite{liu2022online} & $(\epsilon,\delta)$-DP & Objective & $T$ & -- &$\mathcal{O}(\frac{\log(1/\delta)}{\epsilon^2})$  &  Yes\\
\cite{Yuan2023} & $(\epsilon,\delta)$-DP & Output & $T$ & -- &$\mathcal{O}(\frac{\log(1/\delta)}{\epsilon^2})$  &  Yes\\
\cite{cheng2023online} & $\epsilon_{i}$-LDP & Output & $t$ & $\mathcal{O}(\frac{1}{\epsilon^2})$  & --& Yes\\
\cite{Chengradient2023,Chengradienttracking2023} & $\epsilon_{i}$-LDP & Output & $\infty$ & 0  & 0 & No\\
\botrule
\end{tabular*}}{}
\end{table}
\subsection{Other privacy-preserving approaches}
Except for the previously mentioned privacy-preserving methods, various other approaches have been developed to protect the privacy of participating agents' private information in distributed optimization and learning. For example,~\cite{gupta2020optimization} introduced 
a globally balanced correlated perturbation mechanism, employing the Kullback–Leibler divergence for privacy analysis in a statistical sense. ~\cite{gade2020aggregate} and~\cite{Lin2023} developed algorithms that use locally balanced correlated perturbation mechanisms, designed to obscure cost functions and aggregate estimates in distributed aggregative games. However, these balanced correlated perturbation mechanisms require that each agent has a certain number of neighbors who 
do not share information with adversaries. This requirement may not adequately protect the privacy of the agents' private information when all neighboring agents are curious or hostile~\citep{otherreview}. Our recent work~\cite{gaohuan2023} proposed an inherently privacy-preserving approach in the gradient-tracking algorithm, which enables privacy by adding randomness in stepsizes and coupling weights over each iteration. Other inherently privacy-preserving approaches have also been reported, such as the method explored by~\cite{zhang2018privacy}, which enables privacy through function decomposition. Our subsequent efforts have expanded the range of inherently privacy-preserving methods. Specifically,~\cite{Wang2022quantization} focused on the implementation of privacy through stochastic quantization effects and~\cite{uncoordinated2023} employed time-varying heterogeneous stepsizes to ensure privacy. These works collectively contribute to the evolving landscape of privacy preservation in distributed optimization and learning.

\section{Background}\label{section3}
\subsection{Distributed optimization and  learning}\label{section31}
We consider a network consisting of $m$ agents, each of which can exchange information with neighboring agents through a communication graph $\mathcal{G}=([m],\mathcal{E})$, where $[m]=\{1,\cdots,m\}$ denotes the set of agents and $\mathcal{E}\subseteq [m]\times[m]$ denotes the set of edges. An edge $(i,j)\in\mathcal{E}$ represents that agent $j$ can send information to agent $i$. In this case, agent $j$ is called an in-neighbor of agent $i$. We denote the in-neighbor set and the out-neighbor set of agent $i$ as $\mathcal{N}_{i}^{\text{in}}=\{j\in[m]|(i,j)\in\mathcal{E}\}$ and $\mathcal{N}_{i}^{\text{out}}=\{j\in[m]|(j,i)\in\mathcal{E}\}$, respectively. A graph is called undirected if and only if $(i,j)\in\mathcal{E}$ implies $(j,i)\in\mathcal{E}$, and directed otherwise. For a nonnegative weight matrix $W=\{w_{ij}\}\in{\mathbb{R}^{m\times m}}$, we define the induced directed graph as $\mathcal{G}_{W}([m],\mathcal{E}_{W})$, where $w_{ij}>0$ if and only if $(i,j)\in\mathcal{E}_{W}$, and $w_{ij}=0$ otherwise. We let $w_{ii}=-\sum_{j\in{\mathcal{N}_{i}^{\text{in}}}}w_{ij}$ for all $i\in[m]$. Graph $\mathcal{G}_{W}$ is called strongly connected if there exists a directed path between any pair of distinct agents.    

\subsubsection{Distributed offline optimization} 
In distributed optimization and learning, each agent only has access to its local objective function and is limited to communicating with its neighboring agents. This setting requires cooperation among agents to minimize the summation of all individual local objective functions. To formalize this, the optimization problem can be presented in the following general form:
\begin{equation}
\min_{x\in{\mathbb{R}^{n}}}f(x),\quad f(x)=\frac{1}{m}\sum_{i=1}^{m}f_{i}(x),\label{primal}
\end{equation}
where $m$ is the number of agents, $x\in{\mathbb{R}^{n}}$ is a decision variable, and $f_{i}(x):\mathbb{R}^{n}\mapsto \mathbb{R}$ is a local objective function private to agent $i$.  
\subsubsection{Noncooperative game.} 
Considering a noncooperative game among a set of $m$ agents, i.e., $[m]=\{1,\cdots,m\}$, each agent $i,~i\in[m]$ is characterized by a feasible decision set $\Omega_{i}\subseteq \mathbb{R}^{n_{i}}$ and has an objective function $f_{i}(x_{i},\boldsymbol{x}_{-i})$, where $x_{i}\in \Omega_{i}$ is the decision of agent $i$ and $\boldsymbol{x}_{-i}=\text{col}\{x_{1},\cdots,x_{i-1},x_{i+1},\cdots,x_{m}\}$ is the joint decisions of all other agents except agent $i$. Unlike cooperative optimization in~\eqref{primal}, which focuses on a collective goal, in a noncooperative game, each agent only cares about its own interest and aims to minimize its own local objective function. Thus, a normal-form noncooperative game faced by agent $i$ can be formulated as follows:
\begin{equation}
\min_{x_{i}\in \Omega_{i}} f_{i}(x_{i},\boldsymbol{x}_{-i}),\quad\text{s.t}\quad x_{i}\in \Omega_{i}~\text{and}~\boldsymbol{x}_{-i}\in \prod_{j=1,j\neq i}^{m-1}\Omega_{j}. \label{primalgame}
\end{equation}
As a key concept in noncooperative games, Nash equilibrium (NE) is defined as a decision profile where no agent can gain more payoff by unilaterally changing its own decision, provided that the rest of agents keep their
decisions unchanged. This concept has been widely adopted to characterize the outcome of strategic interactions in noncooperative games. To clarify, the formal definition of NE is given below~\citep{ye2021aggregative}:
\begin{definition}[Nash equilibrium]
Nash equilibrium is a decision profile on which no agent can reduce its cost by unilaterally changing its own decision, i.e., a decision profile $\boldsymbol{x}^*=(x_{i}^*,\boldsymbol{x}_{-i}^*)$ is a Nash equilibrium if $f_{i}(x_{i}^*,\boldsymbol{x}_{-i}^*)\leq f_{i}(x_{i},\boldsymbol{x}_{-i}^*),\quad \forall i\in[m].$
\end{definition}
In the full-decision information setting, facilitated by a centralized coordinator, every agent $i$ has access to all other agents' decision variables $x_{-i}$ and can precisely evaluate its own objective function. However, in the partial-decision information setting, where no coordinator exists, each agent  must estimate the actions of all other agents solely based on the messages exchanged with neighboring agents through a communication network. Here, we consider partial-decision information games.

\paragraph{Aggregative games} 
Aggregative game, a subclass of noncooperation game, are played in various practical situations, such as Cournot price, factory production and public good game~\citep{NEsurvey}. Here, we introduce an average stochastic aggregative game. In this setup, each agent $i$ is characterized by a decision set $\Omega_{i}\subseteq \mathbb{R}^{n}$ and has an objective function $f_{i}(x_{i},\bar{x},\xi_{i})$, where $x_{i}$ denotes the decision of agent $i$, $\bar{x}=\frac{1}{m}\sum_{i=1}^{m}x_{i}$ represents the average of all agents' decisions, and $\xi_{i}\in \mathbb{R}^{d}$ is a random vector. Given that each decision variable $x_{i}$ is restricted in $\Omega_{i}$, the average $\bar{x}$ is restricted in $\bar{\Omega}=\frac{1}{m}\sum_{i=1}^{m}\Omega_{i}$~\citep{Aggregative2024}. With this notation, an average stochastic aggregative game that agent $i$ faces can be formulated as follows:
\begin{equation}
\min_{x_{i}\in \Omega_{i}} \mathbb{E}[f_{i}(x_{i},\bar{x},\xi_{i})],\quad\text{s.t}\quad x_{i}\in \Omega_{i}\quad\text{and}\quad\bar{x}\in \bar{\Omega},\label{primalaggregative}
\end{equation}
where the expected value is taken with respect to $\xi_{i}$ and $f_{i}(\cdot)$ and $\Omega_{i}$ are assumed to be known to agent $i$ only.
\subsubsection{Distributed online learning and optimization}
In distributed online learning/optimization, each agent $i,~i\in[m]$ performs learning on sequentially arriving streaming data. More specifically, at each time $t$, agent $i$ acquires a data point $\xi_{i,t}=\{a_{i,t},b_{i,t}\}$, which is independently and identically sampled from an unknown distribution. Using the sample $a_{i,t}$ and the current model parameter $x_{i,t}$, agent $i$ predicts a label $\hat{b}_{i,t}=\langle x_{i,t},a_{i,t}\rangle$, incurring a loss $l(x_{i,t};\xi_{i,t})$ that quantifies the deviation between $\hat{b}_{i,t}$ and the true label $b_{i,t}$. This loss prompts agent $i$ to update its model parameter from $x_{i,t}$ to $x_{i,t+1}$. The objective is to ensure that, based on sequentially acquired data, all agents collectively converge to the same optimal solution $x^*$ to the following stochastic optimization problem:
\begin{equation}
\min_{x_{i}\in{\mathbb{R}^{n}}}f(x),\quad f(x)=\frac{1}{m}\sum_{i=1}^{m}f_{i}(x),\quad f_{i}(x)=\mathbb{E}_{\xi_{i}}[l(x;\xi_{i})].\label{primalonline}
\end{equation}
It can be seen that the local objective function $f_{i}(x)$ is defined as an expectation over random data $\xi_{i}$, which are sampled from an unknown distribution. Since it is inaccessible in practice, an analytical solution to problem~\eqref{primalonline} is unattainable. To address this issue, we focus on solving the following empirical risk minimization problem using sequentially arriving data:
\begin{equation}
\min_{x_{i}\in{\mathbb{R}^{n}}}f_{t}(x),\quad f_{t}(x)=\frac{1}{m}\sum_{i=1}^{m}f_{i,t}(x),\quad f_{i,t}(x)=\frac{1}{t+1}\sum_{k=0}^{t}l(x;\xi_{i,k}),\label{primalonline2}
\end{equation}
where $f_{i,t}(x)$ is determined by the loss function $l(x;\xi_{i,k})$ with $\xi_{i,k}$ representing the $k$-th data sample of agent $i$ at time $k,~k\in[0,t]$. 

\subsection{Differential privacy}\label{section32}
Differential privacy guarantees that the output of computation on a dataset will not significantly change when any single data point in the dataset is changed. This implies that preserving privacy can be seen as equivalent to masking changes in the dataset. To clarify, changes in a dataset are captured by the following concept of adjacency:
\begin{definition}[Adjacency]\label{adjancy}
For two datasets $\mathcal{D}=\{d_{1},\cdots,d_{m}\}$ and $\mathcal{D}'=\{d'_{1},\cdots,d'_{m}\}$, $\mathcal{D}$ and $\mathcal{D}'$ are adjacent if there exists $i\in\{1,\cdots,m\}$ such that $d_{i}\neq d'_{i}$ and $d_{j}={d}'_{j}$ for all $j\neq i$.
\end{definition} 
Definition~\ref{adjancy} introduces a foundational concept of an adjacent relationship, in which two datasets differ by only a single entry while all other entries are identical. In fact, as we will illustrate later, this definition can be extended further to incorporate more complex objects, such as vector norms, datasets of functions and optimization problems. We denote the adjacent relationship between $\mathcal{D}$ and $\mathcal{D}'$ as $\text{Adj}(\mathcal{D},\mathcal{D}')$.

Given a dataset $\mathcal{D}$, we represent a randomized iterative algorithm as a mapping $\mathcal{A}(\mathcal{D}): \mathcal{D}\mapsto O$, where $O$ represents the observation sequence of all shared messages. We define the set of all possible observation sequences as $\mathcal{O}$. 
Then, a randomized iterative algorithm $\mathcal{A}(\cdot)$ that acts on a dataset achieves differentially private if it can ensure that two adjacent datasets are nearly indistinguishable in a probabilistic sense from observing the output of the algorithm $\mathcal{A}(\cdot)$. The formal definition of $\epsilon$-DP is given as follows:
\begin{definition}[$\epsilon$-Differential Privacy]\label{definition9}
For a given $\epsilon\geq0$, a randomized iterative algorithm $\mathcal{A}(\cdot)$ is $\epsilon$-differential privacy if for any two adjacent datasets $\mathcal{D}$ and $\mathcal{D}'$ and the set of all possible observations $\mathcal{O}$, we always have
\begin{equation}
\mathbb{P}\left[\mathcal{A}(\mathcal{D})\in \mathcal{O}\right]\leq e^{\epsilon}\mathbb{P}\left[\mathcal{A}(\mathcal{D}')\in \mathcal{O}\right].
\end{equation}
\end{definition} 

Definition~\ref{definition9} implies that a small change in the dataset will not significantly affect the output of $\mathcal{A}(\cdot)$, thereby ensuring that an adversary cannot distinguish which specific data entry has been changed from 
the output of $\mathcal{A}(\cdot)$ with high probability. The constant $\epsilon$ corresponds to the level of privacy: a smaller $\epsilon$ implies a higher level of privacy.

In certain cases, it is also useful to consider a relaxed notion of $\epsilon$-DP called $(\epsilon,\delta)$-differential privacy, which is defined as follows:
\begin{definition}[$(\epsilon,\delta)$-differential privacy]\label{definition4}
For given $\epsilon\geq0$ and $\delta\geq 0$, a randomized iterative algorithm $\mathcal{A}(\cdot)$ is $(\epsilon,\delta)$-differential privacy if for any two adjacent datasets $\mathcal{D}$ and $\mathcal{D}'$ and the set of all possible observations $\mathcal{O}$, we always have
\begin{equation}
\mathbb{P}\left[\mathcal{A}(\mathcal{D})\in \mathcal{O}\right]\leq e^{\epsilon}\mathbb{P}\left[\mathcal{A}(\mathcal{D}')\in \mathcal{O}\right]+\delta. \label{D4result}
\end{equation}
\end{definition}
It can be seen from Definition~\ref{definition4} that $(\epsilon,\delta)$-differential privacy becomes $\epsilon$-differential privacy when $\delta=0$. The introduction of an additive term $\delta$ in~\eqref{D4result} yields a weaker privacy guarantee than $\epsilon$-differential privacy. This is because besides a small $\epsilon$, there remains a possibility that  $\mathbb{P}\left[\mathcal{A}(\mathcal{D})\in \mathcal{O}\right]$ is larger than $\mathbb{P}\left[\mathcal{A}(\mathcal{D}')\in \mathcal{O}\right]$, potentially revealing whether the input dataset is $\mathcal{D}$ or $\mathcal{D}'$.

Next, we introduce another pivotal concept associated with DP named sensitivity.
\begin{definition}\label{sensitive}
The sensitivity of a randomized iterative algorithm $\mathcal{A}(\cdot)$ is defined to be
\begin{equation}
\Delta_{t}=\sup_{\text{Adj}(\mathcal{D},\mathcal{D}')}\|\mathcal{A}_{t}(\mathcal{D})-\mathcal{A}_{t}(\mathcal{D}')\|_{1},
\end{equation}
where $\mathcal{A}_{t}(\mathcal{D})$ is an implementation of a randomized iterative algorithm $\mathcal{A}(\cdot)$ on the dataset $\mathcal{D}$ and at time $t$.
\end{definition}
The sensitivity in Definition~\ref{sensitive} quantifies the maximum impact that changing a single data entry can have on the algorithm's output. This metric is crucial for determining how much DP-noise (perturbation) required to guarantee a certain privacy-preserving level. Here, we use Laplace noise to enable differential privacy. For a constant $\nu>0$, $\text{Lap}(\nu)$ denotes the Laplace distribution with a probability density function $\frac{1}{2\nu}e^{-\frac{|x|}{\nu}}$. This distribution has a mean of zero and a variance of $2\nu^2$.  Next, we provide the following lemma to characterize the relationship among sensitivity, DP-noise, and the privacy budget: 
\begin{lemma}
~\citep{huang2015} At each iteration $t$, if each agent adds a noise vector $\chi_{t}\in \mathbb{R}^{n}$ consisting of $n$ independent Laplace noises with parameter $\nu_{t}$ such that $\sum_{t=1}^{T}\frac{\delta_{t}}{\nu_{t}}\leq \epsilon$, then the randomized iterative algorithm $\mathcal{A}(\cdot)$ is $\epsilon$-differential privacy for iterations from $t=0$ to $t=T$.
\end{lemma}
\subsubsection{DP in distributed offline optimization}\label{section321}
In distributed offline optimization, each agent's objective function $f_{i}$ contains local and private information, which is only known to agent $i$ and therefore must be kept confidential. Given that the objective functions are the objects whose privacy needs to be protected, the standard definition of adjacency in Definition~\ref{adjancy} thus needs some adjustments. Drawing on insights from~\cite{huang2015} and~\cite{Wangtailoring2023}, let us first characterize a distributed offline optimization $\mathcal{P}$ in~\eqref{primal} by four parameters $(\mathcal{X},\mathcal{F},f,\mathcal{G}_{W}):$
(a) $\mathcal{X}=\mathbb{R}^{n}$ is the domain of optimization; (b) $\mathcal{F}\subseteq\{\mathbb{R}^{n}\mapsto\mathbb{R}\}$ is a set of real-valued and differentiable individual objective functions; (c) $f(x)=\frac{1}{m}\sum_{i=1}^{m}f_{i}(x)$ with $f_{i}(x)\in \mathcal{F}$ for each $i\in[m]$; (d) $\mathcal{G}_{W}$ represents the induced graph by the weight matrix $W$. With this definitions, an adjacent relationship between two optimization problems is defined as follows:
\begin{definition}[Adjacency in distributed offline optimization]\label{Adjacentfunction}
Two distributed optimization problem $\mathcal{P}$ and $\mathcal{P}'$ are adjacency if the following conditions hold:

(i) $\mathcal{X}=\mathcal{X}'$, $\mathcal{F}=\mathcal{F}'$, and $\mathcal{G}_{W}=\mathcal{G}'_{W}$, i.e., the domain of optimization, the set of individual objective functions, and the communication graphs are identical;

(ii) there exists an $i\in[m]$ such that $f_{i}\neq {f}'_{i}$ but $f_{j}={f}'_{j}$ for all $j\in[m],~j\neq i$;

(iii) the different objective functions $f_{i}$ and ${f}'_{i}$ have similar behaviors round $x^*$, which denotes the optimal solution to problem $\mathcal{P}$. More specifically, there exists some $\delta>0$ such that for all $x$ and $x'$ in $B_{\delta}(x^*)\triangleq\left\{x:~x\in R^{n}\big{|}\left\|x-x^*\right\|<\delta\right\}$, we have $\nabla f_{i}(x)=\nabla {f}'_{i}(x')$.
\end{definition}
Definition~\ref{Adjacentfunction} implies that two distributed optimization problems $\mathcal{P}$ and $\mathcal{P}'$ are adjacent if only one agent changes its objective function and all other conditions remain the same. 
\begin{remark}
Definition~\ref{Adjacentfunction}-(ii) permits arbitrary modifications of the objective function from $f_{i}$ to ${f}'_{i}$. However, to guarantee rigorous DP while ensuring provable convergence to an exact optimal solution, such modifications must be constrained. According to Definition~\ref{Adjacentfunction}-(iii), it requires that the gradients $\nabla f_{i}(x)$ and $\nabla {f}'_{i}(x')$ for two adjacent variables, $x$ and $x'$, in the neighborhood of an optimal solution $x^*$, must be identical. Other DP solutions in distributed offline optimization have introduced other limitations, which can be categorized into three classes: (a) all gradients are uniformly bounded, i.e., $\|\nabla f_{i}(x)\|<\delta$ for all $x\in\mathcal{X}$ \citep{huang2015}; (b) the changes of $\nabla f_{i}(\cdot)$ and $\nabla {f}'_{i}(\cdot)$ must be identical, i.e., $\nabla f_{i}(x)-\nabla f_{i}(x')=\nabla {f}'_{i}(x)-\nabla {f}'_{i}(x')$ for all $x,x'\in\mathcal{X}$~\citep{ding2021}; (c) the norm difference between  $\nabla f_{i}(x)$ and $\nabla {f}'_{i}(x)$  are bounded, i.e., $\|\nabla f_{i}(x)-\nabla {f}'_{i}(x)\|<\delta$ for all $x\in{\mathcal{X}}$~\citep{Shi2024}. Definition ~\ref{Adjacentfunction}-(iii) introduces a mind condition and allows more admissible convex functions, such as $f_{i}(x)=ax^{T}x$ and ${f}'_{i}(x)=bx^{T}x$ with $a,b>0$ and $ax=bx'$ for all $x,x'\in B_{\delta}(x^*)$. It is evident that under these functions, conditions (a)-(b) cannot be satisfied.
\end{remark}

Under Definition~\ref{Adjacentfunction}, the mapping $\mathcal{A}(\cdot)$ in Definition~\ref{definition9} takes a distributed offline optimization problem $\mathcal{P}$ or $\mathcal{P}'$ as its argument. In this case, differential privacy ensures that the statistical difference between the outputs of $\mathcal{A}(\mathcal{P})$ and $\mathcal{A}(\mathcal{P}')$ should be (relatively) minimal if the objective function of one agent changes, making it challenging for an adversary observing the output of $\mathcal{A}(\cdot)$ to identify this change.

\subsubsection{DP in noncooperative game}
Following the same statement in Subsubsection~\ref{section321}, we characterize a noncooperative game problem $\mathcal{P}$ in~\eqref{primalgame}  by three parameters $(\Omega,F,\mathcal{G}_{W})$. Here, $\Omega\triangleq \Omega_{1}\times\cdots\times\Omega_{m}$ is the domain of decision variables, $F\triangleq\{f_{1},\cdots,f_{m}\}$ is a set of real-valued and differentiable individual objective functions, and $\mathcal{G}_{W}$ is the communication graph. Subsequently, the adjacency relationship between two games is defined as follows:
\begin{definition}[Adjacency in noncooperative game]\label{definition5}
Two noncooperative games $\mathcal{P}\triangleq(\Omega,F,\mathcal{G}_{W})$ and $\mathcal{P}'\triangleq(\Omega',F',\mathcal{G}'_{W})$ are adjacent if the following conditions hold:

(i) $\Omega=\Omega'$ and $\mathcal{G}_{W}=\mathcal{G}'_{W}$, i.e., the domain of decision variables and the communication graph are identical;

(ii) there exists an $i\in[m]$ such that $f_{i}\neq f'_{i}$ but $f_{j}=f'_{j}$ for all $j\in[m]$ and $j\neq i$.

(iii) the different objective functions $f_{i}$ and $f'_{i}$ have similar behaviors around $\boldsymbol{x}^*$, where $\boldsymbol{x}^*=\text{col}\{x_{1}^*,\cdots,x_{m}^*\}$ denotes a Nash equilibrium to the aggregative game~\eqref{primalaggregative}. More specifically, there exists some $\delta>0$ such that for all $x$ and $x'$ in $B_{\delta}(x^*)\triangleq\left\{x:~x\in R^{n}\big{|}\left\|x-x^*\right\|<\delta\right\}$, we have $\text{Pro}_{\Omega_{i}}[x-\alpha\nabla_{x}f_{i}(x,\cdot)]-x=\text{Pro}_{{\Omega}'_{i}}[x'-\alpha\nabla_{x'}{f}'_{i}(x',\cdot)]-x'$ for all $\alpha>0$, where $\text{Pro}_{\Omega_{i}}[\cdot]$ denotes the Euclidean projection of a vector onto the set $\Omega_{i}$.
\end{definition} 

To ensure rigorous $\epsilon$-DP in distributed NE seeking for a noncooperative game~\eqref{primalaggregative}, an additional condition in Definition~\ref{definition5}-(iii) is required, which is different from~\cite{ye2021aggregative} and~\cite{WangJimin2022} that restrict all pseudo-gradients to be uniformly bounded. In addition, in the absence of set constraints, Definition~\ref{definition5}-(iii) can be simplified to requiring $\nabla_{x}f_{i}(x,\cdot)=\nabla_{x'}{f}'_{i}(x',\cdot)$ for $x$ and $x'$ in the neighborhood of a Nash equilibrium to the game $\mathcal{P}$.

\subsubsection{DP in distributed online learning and optimization}\label{section323}
The standard setting of DP described in Definition~\ref{definition9} assumes that each participating agent contributes a single data point to the input dataset of the algorithm $\mathcal{A}(\cdot)$, and aims to preserve privacy by adding noise to the output in a way that is commensurate with the maximum impact of a single data point.  However, this scenario does not align with many machine learning applications, where each agent contributes a local dataset consisting of multiple data points.  Consequently, most of the current 
DP distributed online optimization/learning results are still restricted by the ``centralized/collective" property of conventional DP framework. They enable DP only when two different ``centralized" datasets $\mathcal{D}$ and $\mathcal{D}'$, which include all agents' data, differ by only one data point while all other data points are identical at each iteration $t$. In such scenarios, the conventional DP framework fails to adequately protect the privacy of each agent's private dataset.  Moreover, the conventional  DP framework requires agents to mutually trust each other to cooperatively determine the DP noise needed to guarantee a global privacy budget $\epsilon=\sum_{i=1}^{m}\epsilon_{i}$ (where $m$ is the number of agents). Thus, it does not explicitly address protection against information inference by participating agents. 

To ensure differential privacy at the agent-level, Local differential privacy (LDP) provides a more user-friendly and stronger privacy protection for distributed optimization and learning. However, in agent-level LDP framework,  the output of the LDP algorithm is required to be insensitive to changes in the local dataset of any agent, rather than to changes in a single data point within a ``centralized" dataset. This alteration significantly increases the challenges associated with LDP-algorithm design.

\paragraph{Local differential privacy}
As an agent-level differential-privacy framework, LDP not only 
prevents external adversaries from extracting raw data through shared information, but it also provides protection against curious neighboring-agents in the network. Before providing the definition of LDP, it is essential to first introduce the concept of adjacency on the local dataset of agent $i$ under sequentially arriving data:
\begin{definition}[Adjacency in LDP-distributed online learning]\label{definition10}
Given two local datasets $\mathcal{D}_{i}=\{\xi_{i,1},\cdots,\xi_{i,T}\}$ and $\mathcal{D}'_{i}=\{{\xi}'_{i,1},\cdots,{\xi}'_{i,T}\}$ for all $i\in[m]$ and any time $T\in \mathbb{N}^{+}$, $\mathcal{D}_{i}$ and ${\mathcal{D}}'_{i}$ are adjacent if there exists a time instant $k\in\{1,\cdots,T\}$ such that $\xi_{i,k}\neq{\xi}'_{i,k}$ while $\xi_{i,t}\neq{\xi}'_{i,t}$ for all $t\neq k,~t\in\{1,\cdots,T\}$.
\end{definition}
According to Definition~\ref{definition10}, two local datasets $\mathcal{D}_{i}$ and $\mathcal{D}'_{i}$ are adjacent if and only if they differ in one entry at some time instant $k$, with all other entries are the same. With this understanding, we are now in a position to formally define LDP as follows:
\begin{definition}[Local differential privacy]\label{definition11}
We say that an implementation $\mathcal{A}_{i}(\cdot)$ of an iterative algorithm $\mathcal{A}(\cdot)$ by agent $i$ provides $\epsilon_{i}$-local differential privacy if for any adjacent datasets $\mathcal{D}_{i}$ and ${\mathcal{D}}'_{i}$, the following inequality holds:
\begin{equation}
\mathbb{P}[\mathcal{A}_{i}(\mathcal{D}_{i},x_{-i})\in \mathcal{O}_{i}]\leq e^{\epsilon}\mathbb{P}[\mathcal{A}_{i}(\mathcal{D}'_{i},x_{-i})\in \mathcal{O}_{i}],\label{D6result}
\end{equation}
where $x_{-i}$ denotes all messages received by agent $i$ and $\mathcal{O}_{i}$ represents the set of all possible observations on agent $i$.
\end{definition}
In Definition~\ref{definition11}, for agent $i$, all received information from neighbors, i.e., $x_{-i}$, is regarded as external information and is beyond its control. This characteristic of the LDP framework removes the need for mutual trust among agents and allows individual agents to choose heterogeneous privacy budgets $\epsilon_{i}$ in a fully distributed manner, thereby making individual agents free to choose desired privacy strengths depending on their practical
needs. Therefore, LDP operates at an agent-level and provides a stronger privacy framework than the conventional ``centralized" DP framework.
\section{DP-Algorithms and Main Results}\label{Section4}
This section reviews existing DP-algorithms and the corresponding results for distributed optimization/learning. Given the vast algorithms in the literature, our focus is on those DP-algorithms that are capable of achieve both optimization accuracy and rigorous DP with a finite privacy budget even in an infinite time horizon.
\subsection{DP-distributed offline optimization algorithms}\label{section41}
We introduce a DP gradient-descent algorithm for undirected graphs, as summarized in Algorithm~\ref{tailoring1}, and a DP gradient-tracking algorithm for general directed graphs, as summarized in Algorithm~\ref{tailoring2}. 
\begin{algorithm}
\caption{DP-oriented static-consensus based distributed optimization (from Algorithm 1 in~\cite{Wangtailoring2023})}
\label{tailoring1} 
\begin{algorithmic}[1]
\STATE {\bfseries Initialization:} Parameters $x_{i,0}\in{\mathbb{R}^{n}}$; nonnegative weight matrix $W$; stepsize $\lambda_{t}$; weakening factor $\gamma_{t}$; Laplace DP-noise  $\chi_{i,t}=\text{col}\{\chi_{i1,t},\cdots,\chi_{in,t}\}$ with $\chi_{ij,t}\sim \text{Lap}(\sigma_{i,t})$.
\FOR {$t=0,1,\cdots,T-1$} 
\STATE Every agent $j$ adds persistent DP-noise $\chi_{j,t}$ to its state $x_{j,t}$, and then sends the obscured state $x_{j,t}+\chi_{j,t}$ to agent $i\in \mathcal{N}_{j}^{\text{out}}$.
\STATE After receiving $x_{j,t}+\chi_{j,t}$ from all $j\in\mathcal{N}_{i}^{\text{in}}$, agent $i$ updates its state as follows:
\STATE $x_{i,t+1}=x_{i,t}+\sum_{j\in{\mathcal{N}_{i}^{\text{in}}}}\gamma_{t}w_{ij}(x_{j,t}+\chi_{j,t}-x_{i,t})-\lambda_{t}\nabla f_{i}(x_{i,t}).$\\
\ENDFOR
\end{algorithmic}
\end{algorithm}
\vspace{-0.5em}

In Algorithm~\ref{tailoring1} (and similar in Algorithm~\ref{tailoring2}), to achieve a strong DP, an independent DP-noise $\chi_{i,t}$ (and for Algorithm~\ref{tailoring2}, additionally $\zeta_{i,t}$) is incorporated into each round of message sharing. This repeated noise-injection will consistently affects the algorithm through inner-agent iterations, leading to a significant reduction in optimization accuracy. To mitigate the influence of persistent DP-noise on the convergence, a decaying sequence $\{\gamma_{t}\}$ (and for Algorithm~\ref{tailoring2}, $\{\gamma_{1,t}\}$ and $\{\gamma_{2,t}\}$) is used. Under some mild assumptions, \cite{Wangtailoring2023} has proved almost sure convergence and $\epsilon$-DP with a finite privacy budget even in the infinite time horizon for Algorithms~\ref{tailoring1} and~\ref{tailoring2}, respectively.
\begin{algorithm}
	\caption{DP-oriented gradient-tacking based distributed optimization (see Algorithm 2 in~\cite{Wangtailoring2023})}
	\label{tailoring2} 
	\begin{algorithmic}[1]
		\STATE {\bfseries Initialization:} Parameters $x_{i,0}\in{\mathbb{R}^{n}}$ and $y_{i,0}=\nabla f_{i}(x_{i,0})$; weight matrices $R$ and $C$; stepsizes $\lambda_{x,t}$ and $\lambda_{y,t}$; weakening factors $\gamma_{1,t}$ and $\gamma_{2,t}$; Laplace DP-noises $\zeta_{i,t}=\text{col}\{\zeta_{i1,t},\cdots,\zeta_{in,t}\}$ with $\zeta_{ij,t}\sim \text{Lap}(\sigma_{\zeta,i,t})$ and $\chi_{i,t}=\text{col}\{\chi_{i1,t},\cdots,\chi_{in,t}\}$ with $\chi_{ij,t}\sim \text{Lap}(\sigma_{\chi,i,t})$.
		\FOR {$t=0,1,\cdots,T-1$} 
		\STATE Every agent $i$ injects zero-mean DP-noises $\zeta_{i,t}$ and $\chi_{i,t}$ to its states $y_{i,t}$ and $x_{i,t}$, respectively.
		\STATE Agent $i$ pushes $R_{ji}(x_{i,t}+\chi_{i,t})$ and $C_{ji}(y_{i,t}+\zeta_{i,t})$ to each agent $j\in \mathcal{N}_{R,i}^{\text{out}}$ and $j\in \mathcal{N}_{C,i}^{\text{out}}$, respectively, and it pulls $R_{ij}(x_{j,t}+\chi_{j,t})$ and $C_{ij}(y_{j,t}+\zeta_{j,t})$ from each $j\in \mathcal{N}_{R,i}^{\text{in}}$ and $j\in \mathcal{N}_{C,i}^{\text{in}}$, respectively. Here, the subscript $R$ or $C$ in neighbor sets indicates the neighbors with respect to the graphs induced by these matrices.
		\STATE agent $i$ chooses $\gamma_{1,t}>0$ and $\gamma_{2,t}>0$ satisfying $1+\gamma_{1,t}R_{ii}>0$ and $1+\gamma_{2,t}C_{ii}>0$ with $R_{ii}=-\sum_{j\in \mathcal{N}_{R,i}^{\text{in}}}R_{ij}$ and $C_{ii}=-\sum_{j\in \mathcal{N}_{C,i}^{\text{out}}}C_{ji}$.
		\STATE Then, agent $i$ updates its state as follows:
		\STATE $x_{i,t+1}=(1+\gamma_{1,t}R_{ii})x_{i,t}+\gamma_{1,t}\sum_{j\in{\mathcal{N}_{R,i}^{\text{in}}}}R_{ij}(x_{j,t}+\chi_{j,t})-\lambda_{x,t}y_{i,t}.$
		\STATE $y_{i,t+1}=(1-\lambda_{y,t}+\gamma_{2,t}C_{ii})y_{i,t}+\gamma_{2,t}\sum_{j\in{\mathcal{N}_{C,i}^{\text{in}}}}C_{ij}(y_{j,t}+\zeta_{j,t})+\nabla f_{i}(x_{i,t+1})-(1-\lambda_{y,t})\nabla f_{i}(x_{i,t}).$
		\ENDFOR
	\end{algorithmic}
\end{algorithm}
\vspace{-0.5em}
\subsection{DP-distributed NE seeking algorithms}\label{section42}
We introduce a DP algorithm for normal-form noncooperative games, as summarized in Algorithm~\ref{NEseeking}, and another for  aggregative games, as summarized in Algorithm~\ref{Aggregative2}.
\vspace{-0.5em}
\begin{algorithm}[H]
\caption{Distributed NE seeking with provable convergence and differential privacy (see Algorithm 1 in~\cite{NEseeking2022})}
\label{NEseeking} 
\begin{algorithmic}[1]
\STATE {\bfseries Initialization:} Stepsizes $\lambda_{t}>0$; weight matrix $W$; weakening factor $\gamma_{t}>0$; 
\STATE Each agent $i$ maintains one decision variable $x_{i,t}^{i}$, and $m-1$ estimates $\boldsymbol{x}_{i,t}^{-i}=\text{col}\{x_{i,t}^{1},\cdots,x_{i,t}^{i-1},x_{i,t}^{i+1},\cdots,x_{i,t}^{m}\}$ of other agents' decision variables. agent $i$ sets $x_{i,0}^{j}$ randomly in $\mathbb{R}^{n_{j}}$ for all $j\in[m].$
\FOR {$t=0,1,\cdots,T-1$} 
\STATE For both its decision variable $x_{j,t}^{j}$ and estimate variables $x_{j,t}^{1},\cdots,x_{j,t}^{j-1},x_{j,t}^{j+1},\cdots,x_{j,t}^{m}$, every agent $j$ adds respective persistent DP noises $\chi_{j,t}^{1},\cdots,\chi_{j,t}^{m}$, and then send the obscured values $x_{j,t}^{1}+\chi_{j,t}^{1},\cdots,x_{j,t}^{m}+\chi_{j,t}^{m}$ to all neighboring agents $i\in{\mathcal{N}_{j}^{\text{out}}}.$
\STATE After receiving $x_{j,t}^{1}+\chi_{j,t}^{1},\cdots,x_{j,t}^{m}+\chi_{j,t}^{m}$ from all neighboring agents $j\in{\mathcal{N}_{i}^{\text{in}}}$, agent $i$ updates its decision and estimate variables:
\STATE $x_{i,t+1}^{i}=x_{i,t}^{i}+\gamma_{t}\sum_{j\in{\mathcal{N}_{i}^{\text{in}}}}w_{ij}(x_{j,t}^{i}+\chi_{j,t}^{i}-x_{i,t}^{i})-\lambda_{t}\nabla_{x_{i}} f_{i}(x_{i,t}^{i},x_{i,t}^{-i}),$
\STATE $x_{i,t+1}^{l}=x_{i,t}^{l}+\gamma_{t}\sum_{j\in{\mathcal{N}_{i}^{\text{in}}}}w_{ij}(x_{j,t}^{l}+\chi_{j,t}^{l}-x_{i,t}^{l}),~\forall l\in[m]~\text{and}~l\neq i.$\\
\ENDFOR
\end{algorithmic}
\end{algorithm}
In Algorithm~\ref{NEseeking}, since $\boldsymbol{x}_{-i,t}$ is not directly available for agent $i$, each agent $i$ generates a local estimate of $\boldsymbol{x}_{t}=(x_{i,t}^{i},\boldsymbol{x}_{i,t}^{-i})$ to approximate all agents' decisions at each iteration $t$.~\cite{NEseeking2022} has shown that Algorithm~\ref{NEseeking} ensures almost sure convergence to the unique NE to problem~\eqref{primalgame} and at the same time preserves rigorous $\epsilon$-DP with a finite cumulative privacy budget even when $T\rightarrow\infty$. 
\vspace{-0.5em}
\begin{algorithm}[H]
\caption{Differentially-private distributed algorithm for stochastic aggregative games with guaranteed convergence (see Algorithm 2 in~\cite{Aggregative2024})}
\label{Aggregative2} 
\begin{algorithmic}[1]
\STATE {\bfseries Initialization:} Stepsizes $\lambda_{x,t}>0$; weight matrix $W$; weakening factor $\gamma_{t}>0$.
\STATE Every agent $i$ maintains one decision variable $x_{i,t}$, which is initialized with a random vector in $\Omega_{i}\subseteq \mathbb{R}^{d}$, and an estimate of the aggregative decision $y_{i,t}$, which is initialized as $y_{i,0}=x_{i,0}$.
\FOR {$t=1,\cdots,T-1$} 
\STATE Every agent $j$ adds persistent DP noise $\zeta_{j,t}$ to its estimate $y_{j,t}$, and then sends the obscured estimate $y_{j,t}+\zeta_{j,t}$ to agent $i\in{\mathcal{N}_{j}^{\text{out}}}$.
\STATE After receiving $y_{j,t}+\zeta_{j,t}$ from all neighboring agents $j\in{\mathcal{N}_{i}^{\text{in}}}$, agent $i$ updates its decision variable and estimate as follows:
\STATE $x_{i,t+1}=\text{Pro}_{\Omega_{i}}[x_{i,t}-\lambda_{t}\tilde{F}_{i}(x_{i,t},y_{i,t},\xi_{i,t})],$ where $\text{Pro}_{\Omega_{i}}$ denotes the Euclidean projection of a vector onto the set $\Omega_{i}$ and $\tilde{F}_{i}(x_{i,t},y_{i,t},\xi_{i,t})$ is given by $\tilde{F}_{i}(x_{i,t},y_{i,t},\xi_{i,t})=\nabla_{x_{i}}f_{i}(x_{i,t},y_{i,t},\xi_{i,t})$.
\STATE $y_{i,t+1}=y_{i,t}+\gamma_{t}\sum_{j\in{\mathcal{N}_{i}^{\text{in}}}}w_{ij}(y_{j,t}+\zeta_{j,t}-y_{i,t}-\zeta_{i,t})+x_{i,t+1}-x_{i,t}.$\\
\ENDFOR
\end{algorithmic}
\end{algorithm}
In Algorithm~\ref{Aggregative2}, to mitigate the influence of noises on the aggregate estimation for $\bar{x}_{t}$, each agent $i$ uses $y_{i,t}+\zeta_{i,t}$ that is shared among its neighbors in its interaction terms $\sum_{j\in{\mathcal{N}_{i}^{\text{in}}}}w_{ij}(y_{j,t}+\zeta_{j,t}-y_{i,t}-\zeta_{i,t})$. Although the noise $\zeta_{i,t}$ for $i=1,\cdots,m$ are independent of all agents, when an agent $i$ has only one neighboring agent $j$, such interaction can lead to a  correlation between the two agents' dynamics. This correlation might allow agent $j$ to infer certain information of agent $i$. This scenario indicates a limitation of the conventional DP framework, which typically relies on a data aggregator to collect data and inject noises. In the distributed setting, this implies an implicit assumption that agents trust each other enough to cooperatively mask shared information to satisfy a common privacy budget. Hence, to completely avoid correlated dynamics among interacting agents,  the LDP framework is presented as a viable solution, with related algorithms to be detailed in the subsequent subsection.

\cite{Aggregative2024} proved that Algorithm~\ref{Aggregative2} converges to the unique NE of the game in~\eqref{primalaggregative} almost surely and achieves $\epsilon$-DP with the cumulative privacy budget is always finite even when the iteration number tends to infinity. 

\subsection{LDP-distributed online learning algorithms}\label{section43}
We introduce an LDP online gradient-descent algorithm for undirected graphs, as summarized in Algorithm~\ref{onlinegradient}, and an LDP online gradient-tracking algorithm for general directed graphs, as summarized in Algorithm~\ref{ogradienttracking}.
\vspace{-0.5em}
\begin{algorithm}[H]
\caption{LDP-distributed online learning for agent $i$ (see Algorithm 1 in~\cite{Chengradient2023})}
\label{onlinegradient} 
\begin{algorithmic}[1]
\STATE {\bfseries Initialization:} Stepsizes $\lambda_{t}=\frac{\lambda_{0}}{(t+1)^{v}}$ with $\lambda_{0}>0$ and $v\in(\frac{1}{2},1)$; weight matrix $W$; weakening factor $\gamma_{t}=\frac{\gamma_{0}}{(t+1)^{u}}$ with $\gamma_{0}>0$ and $u\in(\frac{1}{2},1)$. Random initial decision variable $x_{i,0} \in \Omega$ for all $i\in[m]$.
\FOR {$t=1,\cdots,T-1$} 
\STATE Agent $i$ receives the current data $\xi_{i,t}\in \mathcal{D}_{i,t}$ and sends $x_{i,t}+\chi_{i,t}$ to its neighboring agents $j\in{\mathcal{N}_{i}^{\text{out}}}$.
\STATE By using all available data up to time $t$, i.e., $\xi_{i,k}\in\mathcal{D}_{i,t},~k\in[0,t]$ and the current decision variable $x_{i,t}$, agent $i$ computes the gradient $\nabla f_{i,t}(x_{i,t})=\frac{1}{t+1}\sum_{k=0}^{t}\nabla l(x_{i,t},\xi_{i,k})$.
\STATE After receiving $x_{j,t}+\chi_{j,t}$ from all neighboring agents $j\in{\mathcal{N}_{i}^{\text{in}}}$, agent $i$ updates its decision variable and estimate as follows:
\STATE $x_{i,t+1}=\text{Pro}_{\Omega}[x_{i,t}+\gamma_{t}\sum_{j\in{\mathcal{N}_{i}^{\text{in}}}}w_{ij}(x_{j,t}+\chi_{j,t}-x_{i,t})-\lambda_{t}\nabla f_{i,t}(x_{i,t})],$ where $\text{Pro}_{\Omega}$ denotes the Euclidean projection of a vector onto the set $\Omega$.
\ENDFOR
\end{algorithmic}
\end{algorithm}
\begin{algorithm}
\caption{LDP design for distributed online learning under general directed graphs (see Algorithm 1 in~\cite{Chengradienttracking2023})}
\label{ogradienttracking} 
\begin{algorithmic}[1]
\STATE {\bfseries Initialization:} Stepsizes $\lambda_{t}=\frac{\lambda_{0}}{(t+1)^{v}}$ with $\lambda_{0}>0$ and $v\in(0.5,1)$; weight matrices $R$ and $C$. Randomly initial optimization variables $x_{i,0}\in \mathbb{R}^{n}$, $y_{i,0}\in \mathbb{R}^{n}$, $z_{i,0}=\mathbf{e}_{i}\in \mathbb{R}^{m}$, where $\mathbf{e}_{i}$ has the $i$-th element equal to one and all other elements equal to zero.
\FOR {$t=1,\cdots,T-1$} 
\STATE Using all available data up to time $t$, i.e., $\xi_{i,k}$ for $k\in[0,t]$ and the current decision variable $x_{i,t}$, agent $i$ computes the gradient $\nabla f_{i,t}(x_{i,t})=\frac{1}{t+1}\sum_{k=0}^{t}\nabla l(x_{i,t},\xi_{i,k})$.
\STATE After Pushing $y_{i,t}+\zeta_{i,t}$ to neighbors $j,~j\in{\mathcal{N}_{C,i}^{\text{out}}}$ and pulling $y_{j,t}+\zeta_{j,t}$ from neighbors $j,~j\in{\mathcal{N}_{C,i}^{\text{in}}}$, agent $i$ updates its tacking variable as follows:
\STATE $y_{i,t+1}=(1+C_{ii})y_{i,t}+\sum_{j\in{\mathcal{N}_{C,i}^{\text{in}}}}C_{ij}(y_{j,t}+\zeta_{j,t})+\lambda_{t}\nabla f_{i,t}(x_{i,t})$ with $C_{ii}=-\sum_{j\in \mathcal{N}_{C,i}^{\text{out}}}C_{ji}$.
\STATE After Pushing $x_{i,t}+\chi_{i,t}$ to neighbors $j,~j\in{\mathcal{N}_{R,i}^{\text{out}}}$ and pulling $x_{j,t}+\chi_{j,t}$ from neighbors $j,~j\in{\mathcal{N}_{R,i}^{\text{in}}}$, agent $i$ updates its decision variable and estimate as follows:
\STATE $x_{i,t+1}=(1+R_{ii})x_{i,t}+\sum_{j\in{\mathcal{N}_{R,i}^{\text{in}}}}R_{ij}(x_{j,t}+\chi_{j,t})-\frac{y_{i,t+1}-y_{i,t}}{m[z_{i,t}]_{i}},$ where $[z_{i,t}]_{i}$ denotes the $i$-th element of $z_{i,t}$ and $R_{ii}$ satisfies $R_{ii}=-\sum_{j\in \mathcal{N}_{R,i}^{\text{in}}}R_{ij}$.
\STATE $z_{i,t+1}=z_{i,t}+\sum_{j\in{\mathcal{N}_{R,i}^{\text{in}}}}R_{ij}(z_{j,t}-z_{i,t}).$
\ENDFOR
\end{algorithmic}
\end{algorithm}
\vspace{-0.5em}
By judiciously designing the attenuation sequence $\gamma_{t},$ the stepsize $\lambda_{t},$ and the DP-noise variance sequence $\sigma_{i,t}$,~\cite{Chengradient2023} proved that Algorithm~\ref{onlinegradient} achieves mean square convergence to the optimal solution $x_{t}^*$ to problem~\eqref{primalonline2} and preserves $\epsilon_{i}$-LDP with a finite cumulative privacy budget even when $T\rightarrow\infty$. 

\begin{figure}[H]
\centering
\subfigure[Algorithm~\ref{ogradienttracking},  ``CIFAR 10" dataset, training accuracy]{\includegraphics[height=5.2cm,width=8.1cm]{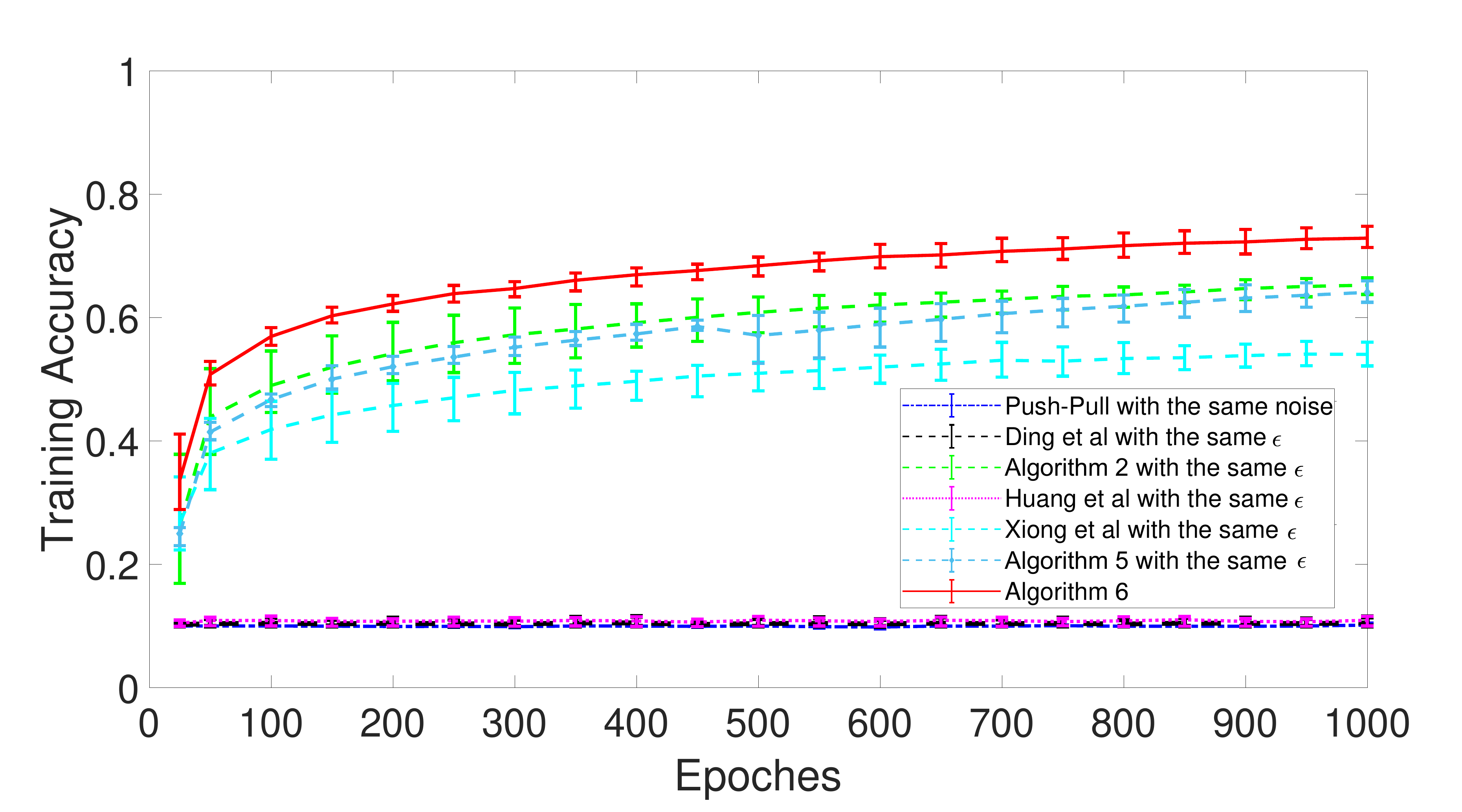}}\quad
\subfigure[Algorithm~\ref{ogradienttracking},``CIFAR 10" dataset, test accuracy]{\includegraphics[height=5.2cm,width=8.1cm]{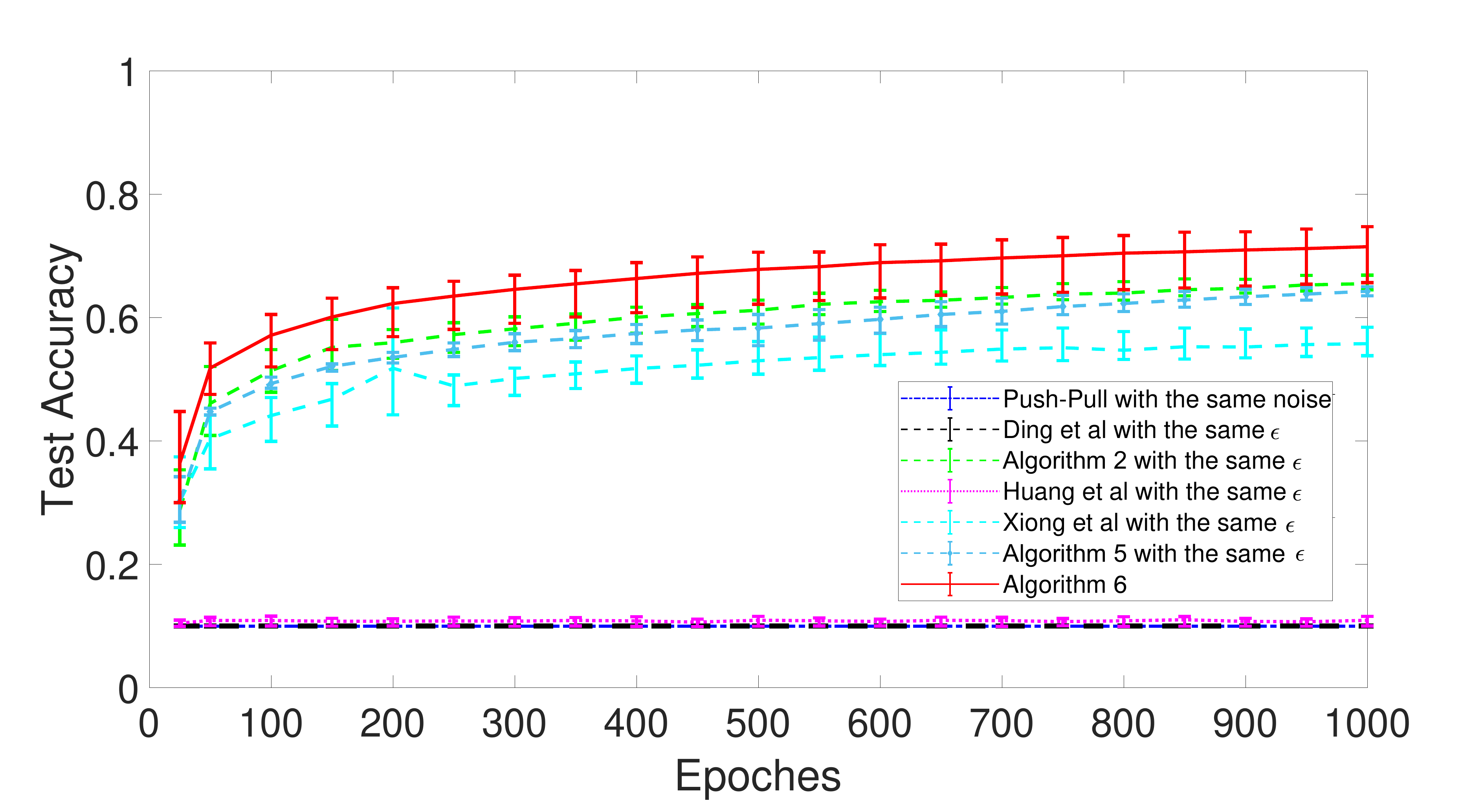}}
\caption{Comparison of Algorithm~\ref{ogradienttracking} with existing DP solutions for distributed learning and optimization, including the DiaDSP algorithm in~\cite{ding2021}, the Algorithm~\ref{tailoring2} from~\cite{Wangtailoring2023}, the DP distributed optimization algorithm in~\cite{huang2015}, the distributed online optimization algorithm in~\cite{xiong2020}, and Algorithm~\ref{onlinegradient} from~\cite{Chengradient2023}. To ensure a fair comparison, the privacy budget for these algorithms is set as the maximum $\epsilon_{i}$ across all agents used in Algorithm~\ref{ogradienttracking}, which corresponds to the weakest level of privacy protection among all agents. Moreover, the conventional Push-Pull gradient-tracking algorithm in~\cite{pushpull} was also evaluated under the same DP noises as those used in Algorithm~\ref{ogradienttracking}.}
\label{onlinecifaralgorithm7}
\end{figure}

In Algorithm~\ref{ogradienttracking}, incorporating the difference $y_{i,t+1}-y_{i,t}$ rather $y_{i,t}$ (which is typically used in conventional gradient-tracking-based algorithm~\citep{pushpull}) into the decision variable update in Line 7 is to resolve the issue of DP-noises accumulation in global gradient estimation. This modification ensures optimization accuracy, as evidenced in Section III-A in~\cite{Chengradienttracking2023}. Moreover, Algorithm~\ref{ogradienttracking} removes the need for a weakening factor in inter-agent iterations, which is crucial in Algorithms~\ref{tailoring1}-\ref{onlinegradient} to simultaneously ensure optimization accuracy and $\epsilon$-DP. Note that this weakening factor reduces the coupling strength among agents, consequently slowing down the  algorithmic convergence speed. Therefore,  Algorithm~\ref{ogradienttracking} is able to achieve faster convergence than Algorithm~\ref{onlinegradient}, as evidenced in Figure~\ref{onlinecifaralgorithm7}. Under some mild assumptions,~\cite{Chengradienttracking2023} has proved that Algorithm~\ref{ogradienttracking} converges in mean square to the optimal solution $x^*$ to problem~\eqref{primalonline} and preserves $\epsilon_{i}$-LDP with a finite cumulative privacy budget even when $T\rightarrow\infty$.

\section{Example Applications}
DP-distributed optimization/learning algorithms can be applied to solve numerous real-world problems, including logistic regression in medical diagnosis~\citep{medicdignosis}, collaborative localization in spectrum sensor networks~\citep{sensor}, demand response in distributed smart grid~\citep{demad}, image classification in distributed deep learning~\citep{classification}, among others. As examples, the following two applications are briefly introduced to illustrate their practicability.
\subsection{Logistic regression}
Logistic regression is a statistical method for analyzing a dataset in which one or more independent variables determine output results. Although originally designed for binary classification tasks, 
logistic regression can be effectively extended to multi-class classification tasks through strategies, such as One-vs-All or One-vs-One. $l_{2}$-logistic regression (ridge regression) is a variation of logistic regression that includes a regularization term. Here, we apply a $l_{2}$-logistic regression model to execute classification tasks on the ``Mushrooms" dataset and the ``Covtype" dataset, respectively.  The loss function is given by 
\begin{equation}
l(x,\xi_{i})=\frac{1}{N_{i}}\sum_{p=1}^{N_{i}}(1-b_{i,p}a_{i,p}^{T}x-\text{log}(s((a_{i,p})^{T}x))+\frac{r_{i}}{2}\|x\|^2,\label{logisticloss}
\end{equation}
where $N_{i}$ is the number of samples, $s(a)$ is the sigmoid function defined as $s(a)=\frac{1}{1+e^{-a}}$, $\xi_{i}=(a_{i,t},b_{i,t})\in \mathcal{D}_{i}$ represents the data point acquired by agent $i$, and $r_{i}>0$ is a regularization parameter proportional to $N_{i}$.

\paragraph{Binary classification on the ``Mushrooms" dataset}
Binary classification on the ``Mushrooms" dataset is a classic task in machine learning, aiming to differentiate between edible and poisonous mushrooms based on various features, such as cap shape, cap color, gill size, and habitat.  Using Algorithm~\ref{onlinegradient} with sequentially arriving data,~\cite{Chengradient2023} trained an $l_{2}$-logistic regression model~\eqref{logisticloss} that achieved high classification accuracy even under the LDP constraints, as evidenced by low training/test losses. This result demonstrates Algorithm~\ref{onlinegradient}'s capability to ensure a good performance while preserving privacy. Similarly, a comparable experiment was conducted in~\cite{Chengradienttracking2023} to evaluate Algorithm~\ref{ogradienttracking}, which yielded comparable results in terms of low training and test losses.

\paragraph{Multi-class classification on the ``Covtype" dataset}
The "Covtype" dataset, also known as the Forest Cover Type dataset, is a widely used dataset in machine learning that aims to predict the forest cover type for $30\times30$ meter cells, based on cartographic variables. This dataset includes seven different forest cover types and is characterized by $54$ features, including soil type, elevation, hillshade, and distance to water features, among others. It represents an example of a multi-class classification task where the goal is to classify each cell into one of the seven forest cover types. For this multi-class classification task,~\cite{Chengradient2023} and~\cite{Chengradienttracking2023} have conducted experiment validation for Algorithm~\ref{onlinegradient} and Algorithm~\ref{ogradienttracking}, respectively. 

\subsection{Convolutional neural network training}\label{section51}
Convolutional Neural Networks (CNNs) stand at the forefront of image classification due to their proficiency in directly processing and learning from image data.  However, as the depth of CNN increases, traditional CNNs usually suffer from the vanishing gradient problem. To solve this issue, ResNet-18 has emerged as an evolutionary development in standard CNN architectures,  aiming to mitigate the vanishing gradient problem in deep learning. In the distributed training of a ResNet-18 architecture, distributed optimization algorithms play a pivotal role in updating the network's weights to minimize the loss function. Here, we introduce the distributed training of a ResNet-18 architecture for image classification tasks on the ``MNIST" dataset and the ``CIFAR-10" dataset, respectively, utilizing categorical cross-entropy loss as the loss function.

\paragraph{Image classification on the ``MNIST" dataset}
The ``MNIST" dataset is a cornerstone in the field of machine learning and computer vision, consisting of $70,000$ handwritten digits ($0$ through $9$). It is typically divided into $60,000$ training images and $10,000$ test images, each a $28x28$ pixel grayscale image. The goal of this classification task is to accurately recognize and classify the handwritten digits into one of the ten possible classes ($0$ through $9$). Given this image classification task,~\cite{Wangtailoring2023} evaluated the performances of Algorithms~\ref{tailoring1} and~\ref{tailoring2}, respectively, even under DP constraints. Moreover, to compare the strength of enabled privacy protection, ~\cite{Wangtailoring2023} also conducted tests by using the DLG attack model proposed in~\cite{DLG}. The training/testing accuracies under different levels of DP-noise and the DLG attacker's inference errors are summarized in Table~\ref{table5}, which shows a trade-off between privacy and accuracy under a fixed iteration number $20,000$.
\paragraph{Image classification on the ``CIFAR-10" dataset}
The ``CIFAR-10" dataset, one of the most widely used datasets in machine learning, presents a greater challenge for training compared to the "MNIST" dataset. It consists of $60,000$ color images across $10$ different classes. The classes represent airplanes, cars, birds, cats, deer, dogs, frogs, horses, ships, and trucks, making it a diverse collection for image classification. The dataset is typically divided into $50,000$ training images and $10,000$ test images. The goal of image classification on the ``CIFAR-10" dataset is to accurately predict an image's category from these ten classes.  In light of this image classification task,~\cite{Chengradient2023} and~\cite{Chengradienttracking2023} evaluated the performances of Algorithms~\ref{onlinegradient} and~\ref{ogradienttracking}, respectively, under LDP constraints. Furthermore, a comparison of  Algorithm~\ref{ogradienttracking} with the Algorithm~\ref{onlinegradient} and other state-of-the-art DP algorithms is summarized in Figure~\ref{onlinecifaralgorithm7}, providing insights into their relative effectiveness.
\section{Future Discussion}
This section aims at pointing out possible future research directions in the area of privacy preservation in distributed optimization/learning.
\subsubparagraph{Performance Improvement}
It can be clearly seen that encryption methods, such as homomorphic encryption and secure multi-party computation protocols, incur significant computational and communication costs. Efforts aimed at reducing these costs could substantially reduce the running time of encryption-based distributed algorithms and expand their applications in large-scale distributed learning. In addition, although there have been results that address the accuracy-privacy dilemma in differential privacy distributed optimization/learning, many of these results sacrifice convergence speed for accuracy and privacy. Minimizing this compromise in convergence speed remains a critical area for future development. 
\subsubparagraph{Inequality constraints}
Addressing coupled inequality constraint has long been an intriguing topic in distributed optimization and learning applications, such as resource allocation in distributed smart grids and robot secure control in distributed wireless networks. However, the study of such problems with privacy-preserving constraints is largely missing. 
\subsubparagraph{Nonconvex objective functions}
Most of the current privacy-preserving results focus on the convex/strongly convex case in distributed optimization and learning. An exception is the recent work~\citep{nonconvex1}, which constructs a Chebyshev polynomial approximation to ensure optimality and leverages the randomness in the blockwise insertions of perturbed vector states for privacy protection. However, in each round of communication among agents, only a portion of the private information is masked, potentially leaving the rest exposed. Hence, the task of privacy-preserving
nonconvex  distributed optimization/learning worths more research efforts.
\subsubparagraph{Nonsmooth objective functions} 
In general, the objective function $f_{i}$ in distributed optimization/learning can be smooth or nonsmooth, particularly in realistic applications involving low-rank, monotonicity, sparsity, and so forth. However, most of the existing privacy-preserving works have focused on distributed optimization/learning with smooth objective functions. Although some works~\citep{nonsmooth1,liu2024} have incorporated DP framework into 
distributed nonsmooth optimization/learning, the accuracy-privacy dilemma still remains unresolved. Therefore, this area of research is still ripe for exploration.

\subsubparagraph{Distributed bilevel optimization}
Bilevel optimization recently has attracted increasing attention due to its great success in solving important machine learning tasks, such as meta learning, reinforcement learning, and hyperparameter optimization. In this respect, a few works have studied privacy-preserving methods for centralized bilevel optimization~\citep{bileveloptimization}. However, there remains a significant gap in research regarding privacy-preserving distributed bilevel optimization and learning.

\section{Conclusion}
This paper has presented a comprehensive survey on privacy-preserving methods for distributed optimization and learning. Specifically, we have reviewed cryptographic methods, differential privacy frameworks, and other approaches that have been used and discussed their advantages and challenges in providing privacy.  Furthermore, we have introduced some differential privacy algorithms that can ensure both privacy and optimization accuracy. A comparison of various works has been conducted, and algorithm implementation in real-world machine learning problems has also been undertaken. Finally, we have presented some directions there are worth exploring. It is our hope that this work will serve as a valuable reference for researchers and practitioners in this specific domain.
\begin{table}[H]
\footnotesize
\TBL{\caption{Training/Test accuracies and DLG attacker's inference errors under differential levels of DP-noise in the image classification experiment by using ``MNIST" dataset (from Table 1 in~\cite{Wangtailoring2023})} \label{table5}}
{\begin{tabular*}{\textwidth}{@{\extracolsep{\fill}}@{}ccccccc@{}}
\toprule
\multicolumn{1}{@{}c}{} 
& \multicolumn{3}{c}{\TCH{Algorithm~\ref{tailoring1}}}
& \multicolumn{3}{c}{\TCH{Algorithm~\ref{tailoring2}}} \\
\colrule
Noise Level\footnotemark{a} & $\times 0.5$ & $\times 1$ & $\times 2$ & $\times 0.5$ & $\times 1$ & $\times 2$\\
\colrule
Training Accuracy & 0.951 & 0.925 & 0.859 & 0.924 & 0.921 & 0.910\\
\colrule
Test Accuracy & 0.951 & 0.929 & 0.861 & 0.926 & 0.922 & 0.913\\
\colrule
Final DLG Error & 310.2 & 350.3 & 412.5 & 301.1 & 336.7 & 389.7\\
\botrule
\end{tabular*}}{\begin{tablenotes}
\footnotetext[a]{Considering the Laplace noise $\text{Lap}(1+0.01t^{0.3})$ as the base level for Algorithm~\ref{tailoring1} and Algorithm~\ref{tailoring2}, respectively.}
\end{tablenotes}}
\end{table}

\bibliographystyle{Harvard}
\bibliography{reference}

\end{document}